\documentclass[12pt,twoside,a4paper]{article}
\usepackage{amsmath,amssymb,latexsym,theorem,epsfig,color,subfigure,footmisc,bbm}
\usepackage[numbers]{natbib}
\usepackage{multirow,graphicx,array,rotating,url,multicol}  
\usepackage{epstopdf,afterpage,wasysym,hyperref}
\usepackage{verbatim,tikz}
\usepackage[normalem]{ulem}
\title{Improving the sharpness in neural network-based parametric post-processing of ensemble forecasts}
\author{\'Agnes Baran and M\'at\'e Mihalina\\ Faculty of Informatics, University of Debrecen, Debrecen, Hungary}

\begin{document}
\maketitle

\begin{abstract}
    Statistical post-processing has proven to be an effective tool in improving  ensemble forecast of different weather variables. Case studies show that post-processing can remedy the typically underdispersive and potentially  biased behaviour of the ensemble while optimizing a proper scoring rule expressing the forecast skill. The price of these positive effects is generally a deterioration in sharpness; the width of the central prediction intervals and the uncertainty of the predictions are increasing, especially for shorter lead times. This work aims to reduce the extent of the latter phenomenon for neural network-based parametric post-processing methods by extending the network's loss function  with a penalty term. We demonstrate the effect of the proposed technique for 2m temperature ensemble forecasts of the European Centre for Medium-Range Weather Forecasts downloaded from the EUPPBench benchmark dataset  and verified against synoptic observations. Here, the predictive distribution is Gaussian, and we use the continuous ranked probability score (CRPS) as loss function. The case studies confirm a substantial relative decrease ($8.2\%-12.5\%$) in the width of the nominal central prediction interval compared to the width of the predictive distribution computed without the penalty term, while there is no deterioration in the mean CRPS of probabilistic forecasts and in the RMSE of the predictive mean. 
\end{abstract}

\section{Introduction}
\label{SecInt}

Accurate and reliable forecasting of different weather variables is of great importance in many areas of industry and economy. In agriculture, forecasting temperature, precipitation, and humidity can help in planning work processes (sowing, irrigation, pest control, harvesting) \cite{agri}, in aviation, the accurate forecasting of visibility and wind speed can be critical for safe transportation \cite{wmo}. Another area that has become increasingly important in recent years is the prediction of weather variables related to renewable energy production. According to the annual report of the International Renewable Energy Agency, in 2025, the renewable energy share of electricity capacity was $49.4\%$, and the role of solar and wind power plants increased drastically  \cite{iea}. In the latter cases, the weather variables affecting production (global radiation and 100m wind speed) are highly volatile, and here accurate forecasts given in small time steps significantly facilitate the estimation of expected production and thus the planning of system operation \cite{baran}. 

To have a sight on the reliability of forecasts, it is important to have some information about their uncertainty.
In response to this demand, almost all meteorological services provide ensemble forecasts in addition to categorical forecasts, where a so-called control term and some interchangeable terms are produced to predict the weather variable of  interest. The forecasts are initialized at the same time (or times) every day for a given time interval, with fixed steps (time horizons). The control term is the result of a single run of a Numerical Weather Prediction (NWP) system, using the best available data and model, while the remaining terms are obtained after slight perturbations of initial conditions and of the model parameters and are considered statistically indistinguishable. For example, the Integrated Forecasting System (IFS) of European Centre for Medium-Range Weather Forecasts (ECMWF) computes a 51-member ensemble for a number of weather variables on a grid covering the entire Earth \cite{ifs}, and, in addition to the IFS, since July 2025,  an Artificial Intelligence Forecasting System (AIFS) \cite{aifs} is in operation for calculation of ensemble forecasts.  The mean of the ensemble typically gives a more accurate forecast than the control term, but the main advantage of the ensemble forecast is that the probabilities of different scenarios can be predicted. Although a smaller ensemble spread means a sharper estimate, experience shows that the raw ensemble is typically underdispersive, meaning that the observations are often above or below the range of the ensemble forecasts, and possibly biased \cite{buizza}. To address this problem, statistical post-processing methods were introduced that proved successful in improving the skill of forecasts (see e.g.,  \cite{hemri} and \cite{vannitsem}). 
The aim of the post-processing is to predict the probability distribution of a weather variable at a given location, time and forecast horizon based on the ensemble forecasts. 
A statistical post-processing method without initial assumptions about the predictive distribution is called nonparametric, see for example, \cite{bremnes} where the probabilistic forecast of precipitation was improved using quantile regression. In contrast to the previous approach,  parametric post-processing methods are where the predictive distribution is sought as a member of a known family of distributions, latter is chosen according to the main characteristics of the weather variable. Here, the parameters of the predictive distribution are estimated based on past forecasts-observation pairs using different techniques, see e.g., \cite{bma} for Bayesian Model Averaging, or \cite{gneiting05} for Ensemble Model Output Statistics (EMOS).  

In the last few years, machine learning-based approaches have become widely applied in this area as well, one of the most intensively used methods is predicting the distribution parameters with a Distributional Regression Network (DRN) \cite{RaspLerch}. There are  a wide range of network architectures and training methods, from simple multilayer perceptrons that use only the ensemble forecasts as input, through networks where the input is augmented  with additional covariates, to more complex multi-step architectures. The common feature of these methods is that the outputs are the parameters of the distribution and the loss function is a proper scoring rule (\cite{gneiting07}) suitable for expressing the improvement of the forecasts, e.g., the continuous ranked probability score (CRPS, \cite{crpsR}). The distribution parameters are predicted by optimizing the given scoring rule, however, other metrics exist to evaluate  the forecast skills; such as sharpness and calibration  (\cite{gneiting07a}). Sharper predictions mean narrower confidence intervals, less uncertainty, but sharpness does not depend on the validating observations, a too narrow confidence interval typically  means a badly calibrated prediction, where the portion of observations falling inside the confidence interval (coverage) is much less than the nominal value corresponding to the given confidence level. As mentioned earlier, the raw ensemble is often underdispersive, having quite narrow prediction intervals, especially for short forecast horizons. While  parametric post-processing methods improve the forecast skill providing a better score and calibration, they significantly widen the prediction intervals.

In the present work, we introduce a penalty technique for DRNs that aims to make sharper predictions while preserving calibration. We augment the loss function (i.e. the CRPS) with a term penalizing wide confidence intervals for a given confidence level, thus forcing the network to find a distribution with a better positioned mean. To validate the effect of the proposed penalty term we apply it for  different network architectures and training methodologies, for 2-meter temperature data downloaded from a benchmark dataset. The case studies confirm that application of the penalty does not only results in a significant  reduction in the width of the confidence intervals, but in a slight improvement of the CRPS and of the categorical forecasts, under a minimal reduction in coverage.   

The rest of the paper is structured as follows;  after a brief introduction to parametric post-processing methods in Sec. \ref{pp}, we define the proposed modification of the loss function in Sec. \ref{newloss}. The dataset used in the case study and the applied models are described in Sec. \ref{SecData} and \ref{Seccompdet}, respectively, while the results and the comparison of the different models are presented in Sec. \ref{Res}.   

\section{Parametric post-processing methods}
\label{pp}

Here, we investigate parametric post-processing methods, where to obtain a probabilistic forecast for the given weather variable, we use a parametric predictive distribution. The type of distribution is chosen based on the main features of the weather variable of interest (e.g., non-negativity, skewness) and on empirical tests performed over different datasets.  
The distribution of temperature observations is typically modeled by a Gaussian distribution (see \cite{gneiting05}), and a truncated normal, a log-normal or a truncated generalized extreme value distribution can be applied for the probabilistic forecast of wind speed (\cite{thor}, \cite{bl}, \cite{tgev}). Total cloud cover forecasts, which are given as the eights of sky covered by clouds, recall for discrete probability distribution \cite{tcc}, while in some cases mixture distributions seem suitable to describe the uncertainty of the forecasts, as  a mixture of a gamma and a truncated normal distribution for visibility forecasts in \cite{baranvis}.

For a given station, time, and forecast horizon, the parameters of the predictive distribution are given as some functions of the ensemble forecasts and possibly other covariates corresponding to the same location and time. The parameters of the link function between the forecasts and the distribution parameters, whether they are the parameters of a linear function (EMOS, \cite{gneiting05}), or the weights of a neural network \cite{RaspLerch}, are computed by optimizing the mean of a proper scoring rule over a training set containing historical forecasts-observation pairs (\cite{gneiting07}). In case of parametric post-processing methods one of the most commonly used scoring rule is the continuous ranked probability score (CRPS, see \cite{crpsR}), which is a negatively oriented score defined as 
\begin{equation}
\label{crps}
\text{CRPS}(F,y)=\int\limits_{-\infty}^\infty (F(z)-{\mathbb I}_{z\geq y})^2dz, 
\end{equation}
where $F$ is the predictive CDF, $y$ is the corresponding observation, while  ${\mathbb I}_{z\geq y}$ is the Heaviside function. 

The CRPS of an ensemble forecast can be obtained by using its empirical CDF in  \eqref{crps}, then (see \cite{enscrps})
\[
\text{CRPS}(\hat F,y) = \frac 1K \sum\limits_{i=1}^K\vert x_i-y\vert -\frac 1{2K^2} \sum\limits_{i=1}^K \sum\limits_{j=1}^K\vert x_i-x_j\vert, 
\]
where $x_1,\dots ,x_K$ is the $K$-member ensemble forecast, $y$ is the observation. 

The CRPS can be interpreted as the generalization of the mean absolute error, it is given in the same units as the observations, and it has a closed form for several distribution families \cite{jordan}. 

\subsection{Distributional regression networks}
In 2018  Rasp and Lerch \cite{RaspLerch} introduced a neural network based technique to estimate the parameters of the predictive distribution, and in the last couple of years it has become a popular and widely applied method. In the case study of the original paper forecasts of the 2m temperature were improved by a feedforward, fully connected multi-layer perceptron, but the method was quickly adapted to the post-processing of other weather variables, moreover different network architectures were introduced (see \cite{ghaz22}, \cite{schultz}, \cite{baran}). The main idea of this method is predicting the distribution parameters via a supervised training of a regression network, where the input is the ensemble forecast corresponding to the given observation, while the loss function is  the mean CRPS of the predictive distribution computed over the training set, provided that it has an analytically closed form.  One reason for the popularity of this technique is that it facilitates the use of various additional input features related to the given weather variable or to the stations, as forecasts of other, relevant  weather variables, station coordinates, etc. However, this flexibility comes with the largest disadvantage of this approach; using a more complex network means a larger number of model parameters   (the weights of the network), which  results in the need of a larger training set.

\subsection{Forecast evaluation}
\label{eval}

In the case studies introduced in Sec. \ref{Res}  the parameters of the predictive distribution $F$ are computed by minimizing the mean CRPS ($\overline{\text{CRPS}}_F$) over the training data, while for the evaluation we calculate the mean over the test data. To have a look at the improvement of the CRPS compared to a reference  $\text{CRPS}_{ref}$, we use the continuous ranked probability skill score (CRPSS), which is a positively oriented score, given as 
\begin{equation}
\label{crpss}
\text{CRPSS}=1-\frac{\overline{\text{CRPS}}_F}{\overline{\text{CRPS}}_{ref}}. 
\end{equation}

The main objective of this work is to produce a  sharper forecast without degrading the CRPS value. The sharpness of a predictive distribution corresponds to the concentration of its PDF, the sharper the distribution, the narrower the prediction intervals. The sharpness is independent of the observations, the aim of the probabilistic forecast is to improve it subject to calibration, where the latter describes the consistency between the predictive distribution and the observations (See \cite{gneiting05}). 

Calibration and sharpness can be investigated with the help of the coverage and average width of the $(100\cdot p)\%$  central prediction interval, where the value of $p$ is typically chosen according to the number of the ensemble members: in the case of a $K$-member ensemble $p=(K-1)/(K+1)$. The coverage of the predictive distribution is the proportion of the observations lying in the given central prediction interval, and it is compared to the coverage of the raw ensemble calculated as the proportion of the observations within the interval spanned by the ensemble members (see \cite{gneiting07}). 

When the  probabilistic forecast of a given weather variable is computed from a probability distribution, the mean of the predictive distribution can be used as a categorical forecast. Then, the accuracy of the categorical forecast can be measured by the root mean square error (RMSE).

\section{Improving the sharpness of the forecast}
\label{newloss}

One of the drivers for using the post-processing methods is the underdispersive nature of the raw ensemble predictions; the observations often fall outside the interval spanned by the ensemble members, resulting in rather poor coverage, especially for short forecast horizons. In the parametric post-processing case studies presented in the literature, it is generally observed that the price of improving the mean CRPS and the coverage is increasing the width of the central prediction interval (\cite{baranvis}, \cite{bl}). The mean CRPS depends on the distribution parameters, i.e., more or less directly on the mean and the standard deviation of the distribution, moreover, the distribution parameters are predicted as a function of the ensemble, but the relationship is quite complex, especially in case of DRNs; the loss function to be optimized has a number of local minimizers. The motivation for this study is to modify the network's loss function by penalizing large variance, thus forcing the optimizer to find a better positioned mean, resulting in a narrower central prediction interval for the same CRPS value. 

The proposed modified CRPS function is given as the sum of the original CRPS and a penalty term;
\begin{equation}
\label{lossC}
\text{CRPS}_{mod}(F,y)=\text{CRPS}(F,y)+ \alpha w(F,p), 
\end{equation}
where $w(F,p)$ is the width of the $(100\cdot p)\%$ central prediction interval of the distribution $F$ predicted by the model, and $\alpha$ is a control parameter. 

We test the effect of the penalization on the ensemble forecasts for 2m temperature, then the predictive distribution is 
a normal distribution having a closed form CRPS formula \cite{RaspLerch}; 
\begin{equation}
\label{loss}
\text{CRPS}({\cal N}(\mu,\sigma),y) = \sigma \left[ \frac{y-\mu}{\sigma}(2\Phi(\frac{y-\mu}{\sigma}) - 1) + 2\varphi(\frac{y-\mu}{\sigma}) - \frac{1}{\sqrt{\pi}} \right],
\end{equation}
where $\Phi(z)$ and $\varphi(z)$ are the CDF and the PDF of the standard normal distribution, respectively. Similarly, $w({\cal N}(\mu,\sigma),p)$ can be expressed in term of the width of the $(100\cdot p)\%$ central prediction interval of the standard normal distribution;
\begin{equation}
\label{penterm}
w({\cal N}(\mu,\sigma),p)= 2\sigma \Phi ^{-1}\left( \frac{1+p}{2}\right).
\end{equation}
The width of the central prediction interval is controlled only by $\sigma$ and not by $\mu$, but while the penalty results in a more concentrated distribution, with lower variance, it also affects the value of $\mu$ through the first term of  \eqref{lossC}.

\section{Data}
\label{SecData}
As mentioned earlier, we demonstrate the advantage of the proposed methodology  for 2-m temperature forecasts. The case studies are based on the EUPPBench (European Postprocessing Benchmark) dataset (\cite{eupp}), which contains forecasts and observations for the years 2017 and 2018. The forecasts are calculated by the ECMFW IFS. ECMWF computes 51-member ensemble forecasts for the entire planet, on a grid resolution of about 25 km, and a deterministic forecast on a finer grid (high-resolution term). The forecasts are initialized every day at 00:00 UTC for a 5-day interval, with a 6-hour time step, i.e. for 20 forecast horizons each time (and a forecast for time horizon 0 is also produced, we remove this value in the case studies). The EUPPBench station-level data includes data for 122 stations from five European countries: Austria, Belgium, France, the Netherlands and Germany. Here the forecasts are forecasts for the grid point (model point) closest to the given station, while the observations are values provided by the national meteorological services. To improve the accuracy of the predictions we extend the input features with the high-resolution forecasts of further weather variables, as total cloud cover (\texttt{tcc}), 10m $U$ and $V$ wind components (\texttt{u10} and \texttt{v10}). Moreover, we compute additional  covariates from the metadata corresponding to the stations and model points and from the temporal data, namely: 
\begin{itemize}
\item \texttt{distance\_2d}: the distance of the station and the corresponding model point obtained from the latitude and longitude values using the Haversine formula (with 6371 km as Earth radius)
\item \texttt{altitude\_diff}: the absolute difference of the station and model point altitude
\item \texttt{lt}: lead time, an integer from range 1-20; the forecast horizon given in hours (6-120) divided by 6
\item \texttt{day\_sin}: 
\begin{equation}\label{daysin}
\texttt{day\_sin} = \sin \left( \frac{\texttt{day\_of\_year} - 1}{365} \cdot \pi \right), 
\end{equation}
where \texttt{day\_of\_year} is the sequence number  of the actual day of the predicted forecast time (in the given year)
\item \texttt{hour\_sin}
\begin{equation}\label{dayhour}
\texttt{hour\_sin} = \sin \left( \frac{\texttt{hour}}{24} \cdot \pi \right), 
\end{equation}
where \texttt{hour} is the hour value of the predicted forecast time within the actual day. 
\end{itemize}

After removing data instances with missing values we obtaine a dataset with 1870260 samples; only data corresponding to the year 2017 are used to train the networks, while the trained models are evaluated on samples from the year 2018.

\section{Models and computational details}
\label{Seccompdet}
All the models we built are feedforward, but not fully connected neural networks with two neurons in the output layer that provide the estimations of the distribution parameters $\mu$ and $\sigma$.  To investigate the effect of the proposed penalty every model is trained with two different losses;  first we use the mean of the CRPS \eqref{loss} as loss function, then we train the model with the modified loss given in  \eqref{lossC}-\eqref{penterm}. 

When training the models, we apply different spatial and temporal resolutions to the data.
For the sake of comparability, we use the 2018 data in all cases to evaluate the performance of the models.

\subsection{Model V1} 
\label{SecV1}
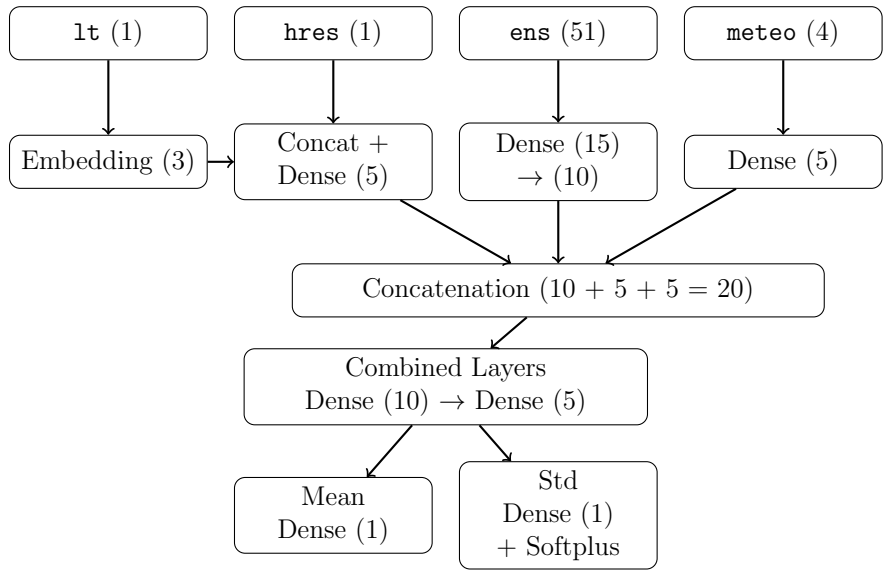
\begin{figure}[h]
\centering
\begin{tikzpicture}[
scale=0.85, transform shape,
block/.style={rectangle, draw, text width=2.8cm, text centered, rounded corners, minimum height=2em},
line/.style={draw, ->, thick}
]
\node [block] (in1) at (0, 0) {\texttt{lt} (1)};
\node [block] (in2) at (3.5, 0) {\texttt{hres} (1)};
\node [block] (in3) at (7, 0) {\texttt{ens} (51)};
\node [block] (in4) at (10.5, 0) {\texttt{meteo} (4)};

\node [block] (emb) at (0, -2) {Embedding (3)};
\node [block] (b2) at (3.5, -2) {Concat + Dense (5)};
\node [block] (b3) at (7, -2) {Dense (15) $\to$ (10)};
\node [block] (b4) at (10.5, -2) {Dense (5)};

\path [line] (in1) -- (emb);
\path [line] (in2) -- (b2);
\path [line] (in3) -- (b3);
\path [line] (in4) -- (b4);
\path [line] (emb) -- (b2); 

\node [block, text width=8cm] (concat) at (7, -4) {Concatenation (10 + 5 + 5 = 20)};
\path [line] (b2) -- (concat);
\path [line] (b3) -- (concat);
\path [line] (b4) -- (concat);

\node [block, text width=6cm] (comb) at (5.25, -5.5) {Combined Layers\\Dense (10) $\to$ Dense (5)};
\path [line] (concat) -- (comb);

\node [block] (out1) at (3.5, -7.5) {Mean\\Dense (1)};
\node [block] (out2) at (7, -7.5) {Std\\Dense (1) + Softplus};
\path [line] (comb) -- (out1);
\path [line] (comb) -- (out2);
\end{tikzpicture}
\caption{Block diagram of the V1 model architecture.}
\label{V1fig}
\end{figure}

The basic model designed for post-processing probabilistic temperature forecasting is a multi-branch, feedforward deep neural network. The architecture transforms different types of inputs through parallel branches, as illustrated in Fig. \ref{V1fig};  
\begin{itemize}
\item[(1)] Lead time embedding: The variable \texttt{lt} computed as the forecast time horizon given in hours divided by six (resulting in integers between 1 and 20) is mapped by an embedding layer into a 3-dimensional dense vector representation.
\item[(2)] High-resolution branch: The high-resolution temperature forecast is concatenated with the embedded time step, and the vector is then further transformed by a 5-neuron dense layer with ReLU activation.
\item[(3)] Ensemble branch: The 51-member ensemble is transformed by two densely connected
hidden layers (with 15 and then 10 hidden units) and ReLU activation functions, extracting higher-level representations from the raw data.
\item[(4)] Additional meteorological variables branch: total cloud cover, wind components and the \texttt{day\_sin} variable calculated with formula \eqref{daysin} are entered into the model through a layer consisting of 5 neurons, transformed with a ReLU activation function.

\end{itemize}

The outputs of the parallel branches are concatenated, followed by a final block consisting of two hidden layers (10 and 5 neurons, ReLU). The architecture ends in two parallel output layers: a layer with linear activation estimates the mean of the distribution, while a layer with a Softplus activation function provides the estimation of the standard deviation, guaranteeing its strict positivity. The network weights are initialized from a normal distribution ($\mu = 0$, $\sigma = 0.01$), while the bias parameters are set to zero.

We apply the model as global model; a single network is trained managing all lead times and stations together, using all data from 2017 as a training set.

\subsection{Model V2}
\label{SecV2}
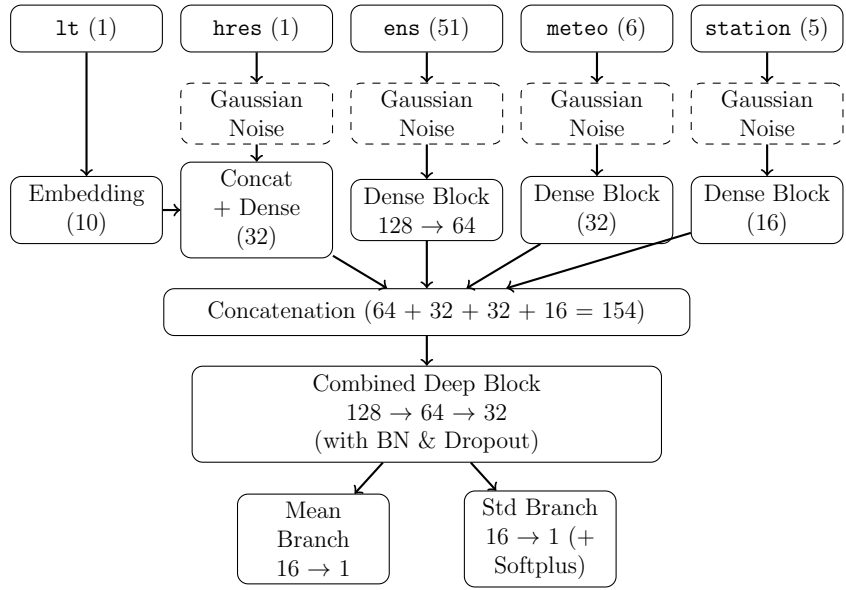
\begin{figure}[h]
\centering
\begin{tikzpicture}[
scale=0.75, transform shape,
block/.style={rectangle, draw, text width=2.4cm, text centered, rounded corners, minimum height=2em},
noise/.style={rectangle, draw, text width=2.4cm, text centered, rounded corners, minimum height=1.5em, dashed},
line/.style={draw, ->, thick}
]
\node [block] (in1) at (0, 0) {\texttt{lt} (1)};
\node [block] (in2) at (3.0, 0) {\texttt{hres} (1)};
\node [block] (in3) at (6.0, 0) {\texttt{ens} (51)};
\node [block] (in4) at (9.0, 0) {\texttt{meteo} (6)};
\node [block] (in5) at (12.0, 0) {\texttt{station} (5)};

\node [noise] (n2) at (3.0, -1.5) {Gaussian Noise};
\node [noise] (n3) at (6.0, -1.5) {Gaussian Noise};
\node [noise] (n4) at (9.0, -1.5) {Gaussian Noise};
\node [noise] (n5) at (12.0, -1.5) {Gaussian Noise};

\path [line] (in2) -- (n2);
\path [line] (in3) -- (n3);
\path [line] (in4) -- (n4);
\path [line] (in5) -- (n5);

\node [block] (emb) at (0, -3.2) {Embedding (10)};
\node [block] (b2) at (3.0, -3.2) {Concat + Dense\\(32)};
\node [block] (b3) at (6.0, -3.2) {Dense Block\\128 $\to$ 64};
\node [block] (b4) at (9.0, -3.2) {Dense Block\\(32)};
\node [block] (b5) at (12.0, -3.2) {Dense Block\\(16)};

\path [line] (in1) -- (emb);
\path [line] (n2) -- (b2);
\path [line] (n3) -- (b3);
\path [line] (n4) -- (b4);
\path [line] (n5) -- (b5);
\path [line] (emb) -- (b2);

\node [block, text width=9cm] (concat) at (6.0, -5) {Concatenation (64 + 32 + 32 + 16 = 154)};
\path [line] (b2) -- (concat);
\path [line] (b3) -- (concat);
\path [line] (b4) -- (concat);
\path [line] (b5) -- (concat);

\node [block, text width=8cm] (comb) at (6.0, -6.8) {Combined Deep Block\\128 $\to$ 64 $\to$ 32\\(with BN \& Dropout)};
\path [line] (concat) -- (comb);

\node [block] (out1) at (4.0, -9) {Mean Branch\\16 $\to$ 1};
\node [block] (out2) at (8.0, -9) {Std Branch\\16 $\to$ 1 (+ Softplus)};
\path [line] (comb) -- (out1);
\path [line] (comb) -- (out2);
\end{tikzpicture}
\caption{Block diagram of the V2 model architecture.}
\label{V2fig}
\end{figure}

The V2 model is an extended variant of the V1 architecture (see Fig. \ref{V2fig}). We expand the input features, add a new branch to take into account the spatial location of the stations and model points, and in order to strengthen the temporal cyclicity, we also use the \texttt{hour\_sin} variable defined by formula \eqref{dayhour} as an input to the meteo branch.
To increase the stability of the training process and reduce overfitting, we add a GaussianNoise ($\sigma = 0.01$) module to the input layers, except the lead time input, this regularization step injects noise into the input data only during the training phase.
The dimensionality of the lead time embedding has been increased to 10 in this version.
The complexity of the parallel feature extraction branches increase significantly, we integrate Batch Normalization and Dropout layers for regularization.

The modified and the new input braches: 
\begin{itemize}
\item[(4)] Meteorological variables branch: total cloud cover, wind components,  the \texttt{day\_sin} and \texttt{day\_hour} variables, moreover the width of the raw ensemble.
\item[(5)] Station coordinates branch: the spatial coordinates of the stations (latitude, longitude, altitude), moreover  the height difference (\texttt{altitude\_diff}) and distance (\texttt{distance\_2d}) defined in Section \ref{SecData}.
\end{itemize}

Like V1, model V2 is trained as a global model. 

\subsection{Model V2\_R}
\label{SecRV}

\begin{figure}[h]
\centering
\begin{tikzpicture}[
scale=0.75, transform shape,
block/.style={rectangle, draw, text width=2.4cm, text centered, rounded corners, minimum height=2em},
noise/.style={rectangle, draw, text width=2.4cm, text centered, rounded corners, minimum height=1.5em, dashed},
line/.style={draw, ->, thick}
]
\node [block] (in1) at (0, 0) {\texttt{lt} (1)};
\node [block] (in2) at (3.0, 0) {\texttt{hres} (1)};
\node [block] (in3) at (6.0, 0) {\texttt{ens} (51)};
\node [block] (in4) at (9.0, 0) {\texttt{meteo} (6)};
\node [block] (in5) at (12.0, 0) {\texttt{station} (5)};

\node [noise] (n2) at (3.0, -1.5) {Gaussian Noise};
\node [noise] (n3) at (6.0, -1.5) {Gaussian Noise};
\node [noise] (n4) at (9.0, -1.5) {Gaussian Noise};
\node [noise] (n5) at (12.0, -1.5) {Gaussian Noise};

\path [line] (in2) -- (n2);
\path [line] (in3) -- (n3);
\path [line] (in4) -- (n4);
\path [line] (in5) -- (n5);

\node [block] (emb) at (0, -3.2) {Embedding (6)};
\node [block] (b2) at (3.0, -3.2) {Concat + Dense\\(8)};
\node [block] (b3) at (6.0, -3.2) {Dense Block\\32 $\to$ 16 \\ (with 0.2 Dropout)};
\node [block] (b4) at (9.0, -3.2) {Dense Block\\(8)};
\node [block] (b5) at (12.0, -3.2) {Dense Block\\(8)};

\path [line] (in1) -- (emb);
\path [line] (n2) -- (b2);
\path [line] (n3) -- (b3);
\path [line] (n4) -- (b4);
\path [line] (n5) -- (b5);
\path [line] (emb) -- (b2);

\node [block, text width=9cm] (concat) at (6.0, -5) {Concatenation (16 + 8 + 8 + 8 = 40)};
\path [line] (b2) -- (concat);
\path [line] (b3) -- (concat);
\path [line] (b4) -- (concat);
\path [line] (b5) -- (concat);

\node [block, text width=8cm] (comb) at (6.0, -6.8) {Combined Deep Block\\40 $\to$ 32 $\to$ 16\\(with BN \& 0.2 Dropout)};
\path [line] (concat) -- (comb);

\node [block] (out1) at (4.0, -9) {Mean Branch\\16 $\to$ 1};
\node [block] (out2) at (8.0, -9) {Std Branch\\16 $\to$ 1 (+ Softplus)};
\path [line] (comb) -- (out1);
\path [line] (comb) -- (out2);
\end{tikzpicture}
\caption{The block diagram of the  V2{\_}R model.}
\label{V2Rfig}
\end{figure}
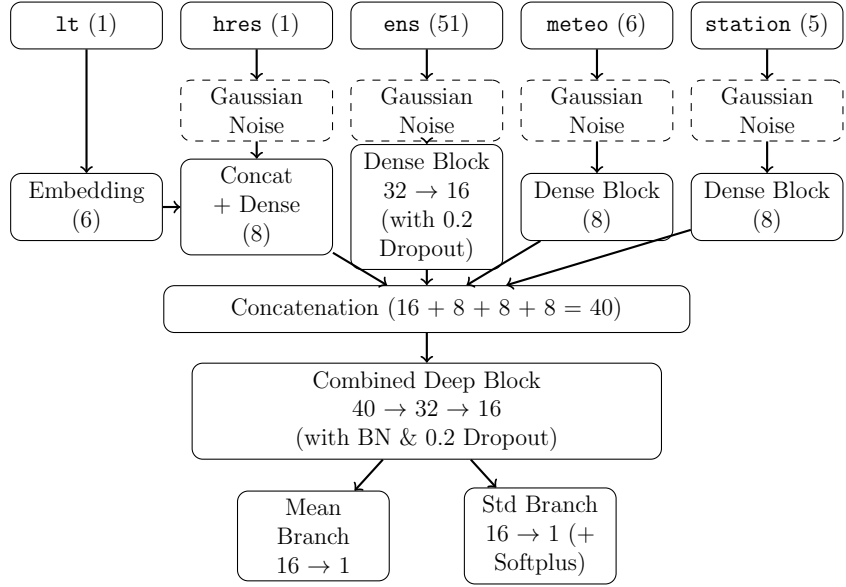

A possible temporal resolution of the data and training is the rolling window training, which is one of the most widely used methods of statistical post-processing. Then, the training set is not a static period  closed in the past, independent of the forecast initialization date, but a fixed-length moving time window containing data for a given number of days prior to the date of initialization. Here, based on empirical tests, we use a 40-day rolling window; a separate model is trained for each day of  2018, always using  the latest available 40-day historical data set as training set. In the rolling training window approach, slightly less training data is available in each iteration, so to avoid overfitting we use a smaller version of the V2 model, see Fig. \ref{V2Rfig}. 

\subsection{Model V2\_LT}
\label{SecLT}

In the  trainings described earlier, we treat all lead times together, modeling the dependence on the forecast horizons by using the \texttt{lt} variable as an input of the network. The disadvantage of this approach is that the inaccuracy of the NWP model and the uncertainty of the ensemble forecasts increase as we move away from the time of the forecast initialization, which is not necessarily compensated by the \texttt{lt} value given as an input feature. 
In a lead time based training, a separate network is trained for each lead time, we applied this technique for the training of the model V\_LT. The model has the same architecture as V2, only the \texttt{lt} and \texttt{hres} input branches are missing, but the \texttt{ens} branch is extended with the high-resolution term (meaning 52 input features here). 
The application of this approach is usually hampered by insufficient amount of data, as here we can use about one twentieth of the training dataset (data corresponding to the given lead time) to train each model. In our case, the amount of available data made it possible to apply the method.

\subsection{Model V2\_L}
\label{SecLoc}

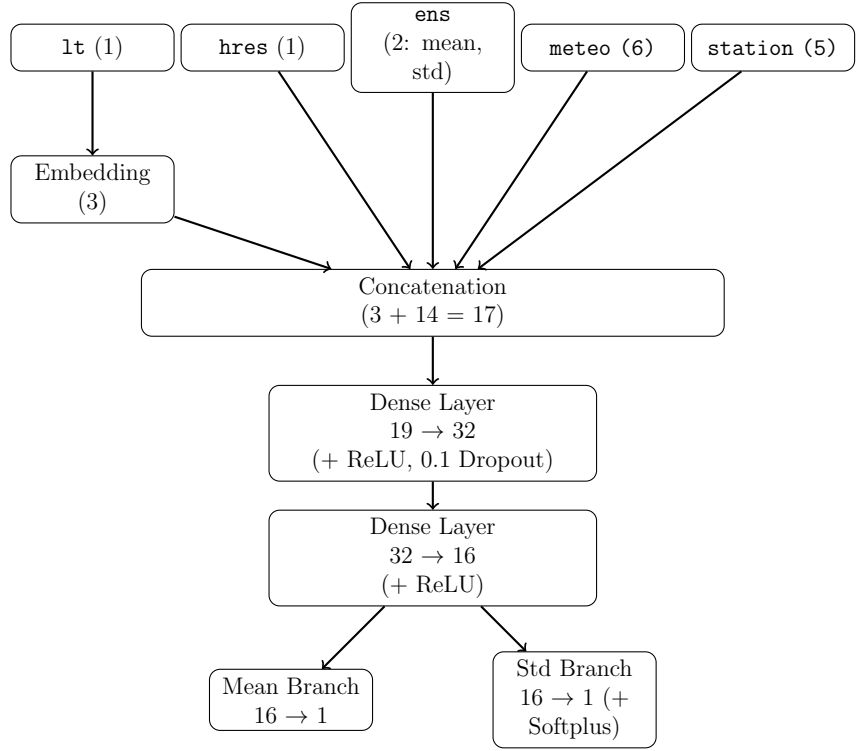
\begin{figure}[ht]
\centering
\begin{tikzpicture}[
scale=0.75, transform shape,
block/.style={rectangle, draw, text width=2.6cm, text centered, rounded corners, minimum height=2em},
line/.style={draw, ->, thick}
]
\node [block] (in1) at (0, 0) {\texttt{lt} (1)};
\node [block] (in2) at (3.0, 0) {\texttt{hres} (1)};
\node [block] (in3) at (6.0, 0) {\texttt{ens}\\(2: mean, std)};
\node [block] (in4) at (9.0, 0) {\texttt{meteo (6)}};
\node [block] (in5) at (12.0, 0) {\texttt{station (5)}};

\node [block] (emb) at (0, -2.5) {Embedding (3)};
\path [line] (in1) -- (emb);

\node [block, text width=10cm] (concat) at (6.0, -4.5) {Concatenation\\(3 + 14 = 17)};
\path [line] (emb) -- (concat);
\path [line] (in2) -- (concat);
\path [line] (in3) -- (concat);
\path [line] (in4) -- (concat);
\path [line] (in5) -- (concat);

\node [block, text width=5.5cm] (trunk1) at (6.0, -6.8) {Dense Layer\\19 $\to$ 32\\(+ ReLU, 0.1 Dropout)};
\path [line] (concat) -- (trunk1);

\node [block, text width=5.5cm] (trunk2) at (6.0, -9.0) {Dense Layer\\32 $\to$ 16\\(+ ReLU)};
\path [line] (trunk1) -- (trunk2);

\node [block] (out1) at (3.5, -11.5) {Mean Branch\\16 $\to$ 1};
\node [block] (out2) at (8.5, -11.5) {Std Branch\\16 $\to$ 1 (+ Softplus)};
\path [line] (trunk2) -- (out1);
\path [line] (trunk2) -- (out2);
\end{tikzpicture}
\caption{The block diagram of the V2\_L model.}
\label{V2Lfig}
\end{figure}

In the last case study we apply a spatial decomposition for the data. In contrast to global training, during local modeling, a separate, unique model is fitted to each station, which uses only the data of the given location. This allows to take into account the specific topographic and microclimatic characteristics of the stations. 
Since these models are optimized on spatially isolated datasets from individual meteorological stations, the size of the training set is drastically reduced, requiring the introduction of an even more simplified architecture (see Fig. \ref{V2Lfig}). In addition to the reduction in the number of neurons in each layer, we would like to draw attention to the change in the input branch of the ensemble variables; instead of the 51-member ensemble the network only receives its mean and standard deviation as input variables.

\subsection{Computational details}
\label{compdet}

The presented models were implemented in the PyTorch framework.  The AdamW optimizer with an adaptive weight decay of $10^{-4}$ was used to train the models. To speed up training and optimize memory usage, we used mixed precision training (torch.amp.autocast) and hardware-specific accelerations. To prevent gradient explosions, we also introduced gradient clipping with a maximum norm of 1.0. 

To avoid overfitting, we used EarlyStopping as a stopping criterion;
$20\%$ of the training data (randomly selected) was set aside as a validation set.
During training, we did not take this data into account when updating the network weights,
but we also monitored the loss function on this set at the end of each epoch.
The training was stopped when the loss measured on the validation set did not improve for 3-7 consecutive epochs (patience), depending on the configuration.
This was supplemented by a learning rate scheduler (ReduceLROnPlateau), which halved the learning rate (up to a minimum of $10^{-5}$) in case the validation error stagnated for 2 epochs. In order to ensure training stability, we also implemented an additional “Soft/Hard Reset” mechanism: if the model did not reach the target error limit within a specified  period, the learning rate was automatically reset to its initial value, and in the extreme case, the model’s weights were completely reset (Hard Restart).

For models with constrained loss functions, the penalty parameter ($\alpha$) was set based on pre-testing with the V2 model. During the experiments, $\alpha = 0.006$ proved to be optimal, as it allowed for the largest reduction in the width of the central prediction interval with minimal degradation of the calibration. The $p$ parameter of the penalty term \eqref{penterm} was chosen according to the number of the ensemble terms ($K=51$); $p=\frac{K-1}{K+1}=\frac{50}{52}$. These values were used consistently in all subsequent experiments; the specific hyperparameters for each training strategy are summarized in Table \ref{tab:configs}

\begin{table}[h]
    \centering{\small
    \renewcommand{\arraystretch}{1.2}
    \begin{tabular}{|l|l|c|c|c|}
        \hline
        \textbf{Training} & \textbf{Model} & \textbf{Batch-size} & \textbf{Learning rate} & \textbf{Patience} \\
        \hline
        Global & \texttt{V1}, \texttt{V2}  & $16\,384$ & $10^{-3}$ & $7$ \\
        \hline
        Lead time based & \texttt{V2\_LT} & $1\,024$ & $10^{-3}$ & $6$ \\
        \hline
        Rolling training window & \texttt{V2\_R} & $8\,192$ & $7 \cdot 10^{-4}$ & $3$ \\
        \hline
        Local & \texttt{V2\_L} & $64$ & $10^{-3}$ & $6$ \\
        \hline
    \end{tabular}}
     \caption{The hyperparameters of the different training configurations}
    \label{tab:configs}
\end{table}

To ensure the stability of the predictions, the models were trained multiple times for each configuration (10 times in the global case, and 5 times in all other cases), and the distribution parameters obtained for each input were calculated as the average of the predictions of these networks for the same input. 

For each model, standardization was applied to the numerical predictors (deterministic values, wind components, total cloud cover, ensemble terms, height difference, and distance) to achieve zero mean and unit standard deviation (StandardScaler).

\section{Results}
\label{Res} 

Each model was trained with two different loss functions; first we used the mean CRPS without the introduced penalty term, we will refer these cases as the unconstrained models, then the models were retrained with the penalized loss function, these models are called constrained models in the analysis of the results. The constrained and unconstrained model versions differed only in the loss functions, all  hyperparameters were the same. The goal of the new loss function was to improve the sharpness of the probabilistic forecasts by reducing the width of the $(100\cdot p)\%$ central prediction intervals while preserving the calibration of the forecasts. The $p$ value is a parameter of the penalty function, we chose $p=\frac{50}{52}$, see Sec. \ref{eval}.

\begin{table}[htbp]
\centering{\small 
\begin{tabular}{l|cc|cc}
\hline
&\textbf{mean}&\textbf{relative}&\textbf{mean}&\textbf{relative}\\
 \textbf{Model} & \textbf{width} & \textbf{change} & \textbf{coverage} & \textbf{change}\\ \hline\hline
 raw ensemble & $4.36$ && $69.57\%$ &  \\
 \hline 
 V1 constrained & $6.11$ & $-12.46\%$ & $91.36\%$&$-2.35\%$\\
 V1 unconstrained& $6.98$ &&$93.56\%$\\ 
 \hline
  V2 constrained & $5.96$ & $-10.13\%$ &$92.23\%$&$-2.26\%$\\
  V2 unconstrained&$6.61$ &&$94.36\%$\\
  \hline
  V2\_LT constrained & $6.09$ & $-8.20\%$ &$92.19\%$&$-2.04\%$ \\
  V2\_LT unconstrained& $6.63$ &&$94.11\%$ \\
  \hline 
  V2\_R constrained& $7.08$ & $-9.98\%$ &$94.50\%$&$-1.80\%$\\
V2\_R unconstrained& $7.87$ &&$96.23\%$ \\ 
\hline 
  V2\_L constrained & $6.00$ & $-8.59\%$ &$91.70\%$&$-2.05\%$\\
  V2\_L unconstrained& $6.56$ &&$93.62\%$\\ \hline
\end{tabular}}
\caption{Average width and coverage values on the test set for the different models}
\label{tab:coverage}
\end{table}

As shown in Table \ref{tab:coverage} the width of the central prediction interval is  significantly smaller for the models with the penalized loss than for the models with the original loss functions. This is accompanied by a decrease of the coverage values, but the relative deterioration is much smaller. However, there are no degradation in the mean CRPS and the RMSE values, and, in fact a slight improvement can be observed on the test set in all cases, see Table  \ref{tab:crps}. To compute the RMSE   the mean of the predicted distribution was used as a categorical prediction. 
For comparison, we also report the values corresponding to the raw ensemble forecasts. 

To better understand the reason for the improving CRPS and RMSE values, we examined the changes in the distribution parameters. Not surprisingly, $\sigma$ (standard deviation) decreased significantly in all cases, but at the same time the values of $\mu$ (mean parameter) were also shifted,  presumably closer to the observations, the average changes are reported in Table \ref{tab:params}. Here, the differences were calculated by subtracting the distribution parameters obtained by the penalized model from the parameters provided by the original model, and in case of the $\mu$ parameters we took the absolute values of the results. 

\begin{table}[h]
\centering{\small 
\begin{tabular}{lcccc}
\hline
&\multicolumn{2}{c}{\bf Training set}&\multicolumn{2}{c}{\bf Test set}\\
\hline
 \textbf{Model} & \textbf{mean CRPS} & \textbf{RMSE} & \textbf{mean CRPS} & \textbf{RMSE}\\ \hline\hline
 raw ensemble & $1.1749$ & $2.0593$ & $1.1981$ & $2.0814$ \\
 \hline 
 V1  constrained& $0.9171$ & $1.7219$& $0.9931$ & $1.8466$ \\
 V1 unconstrained& $0.9445$ & $1.7184$& $1.0198$ & $1.8479$ \\ 
 \hline
  V2 constrained & $0.8411$ & $1.5811$ & $0.9429$ & $1.7569$\\
  V2 unconstrained& $0.8558$ & $1.5949$ & $0.9485$ & $1.7578$\\
  \hline
  V2\_LT constrained & $0.8247$ & $1.5567$ & $0.9642$ & $1.8114$\\
  V2\_LT unconstrained& $0.8244$ & $1.5522$& $0.9645$ & $1.8129$ \\
  \hline 
  V2\_R constrained& $0.9440$ & $1.7387$ & $1.0036$ & $1.9032$\\
V2\_R unconstrained & $0.9511$ & $1.7429$ & $1.0092$ & $1.9571$\\ 
\hline 
  V2\_L constrained & $0.8868$ & $1.6474$ & $0.9563$ & $1.7652$\\
  V2\_L unconstrained & $0.8910$ & $1.6523$ & $0.9570$ & $1.7694$\\ \hline

\end{tabular}}
\caption{Mean CRPS and RMSE values for the different models}
\label{tab:crps}
\end{table}

\begin{table}[h]
\centering{\small 
\begin{tabular}{lcc}
\hline
 \textbf{Model} & \textbf{$\mu$ abs difference} & \textbf{$\sigma$ difference} \\ \hline\hline
 V1 constrained & 0.103055& $-0.144858$\\
 \hline
  V2 constrained & 0.176803&$-0.172969$\\
  \hline
  V2\_LT constrained & 0.209424&$-0.131193$\\
  \hline 
  V2\_R constrained&0.250466&$-0.189722$ \\
\hline 
  V2\_L constrained &0.181493&$-0.136073$\\
 \hline
\end{tabular}}
\caption{Average absolute change of the $\mu$ parameters and average change of the $\sigma$ values on the test set with respect to the corresponding unconstrained model. }
\label{tab:params}
\end{table}

\begin{figure}[h]
    \centering
    \includegraphics[width=0.48\textwidth]{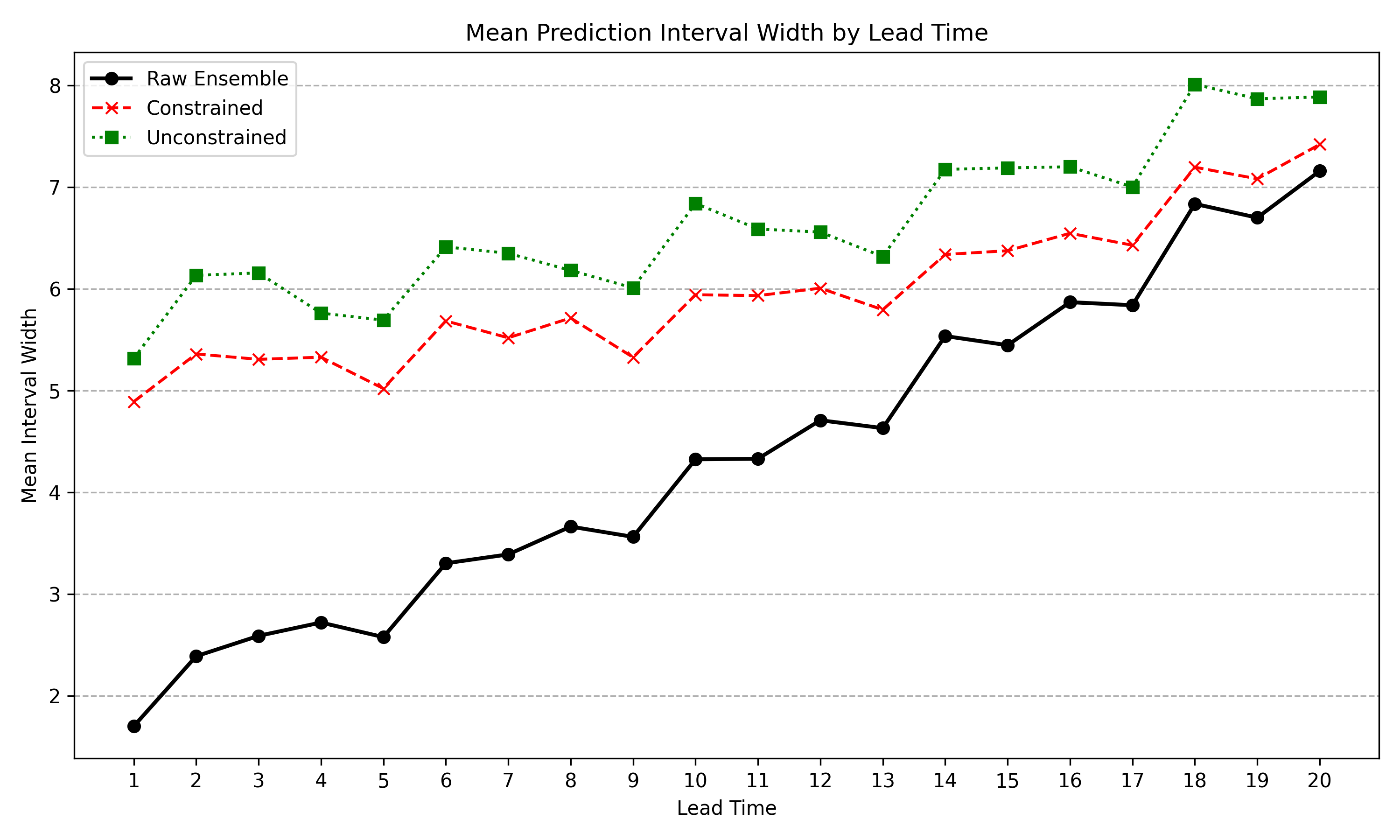}
    \hfill
    \includegraphics[width=0.48\textwidth]{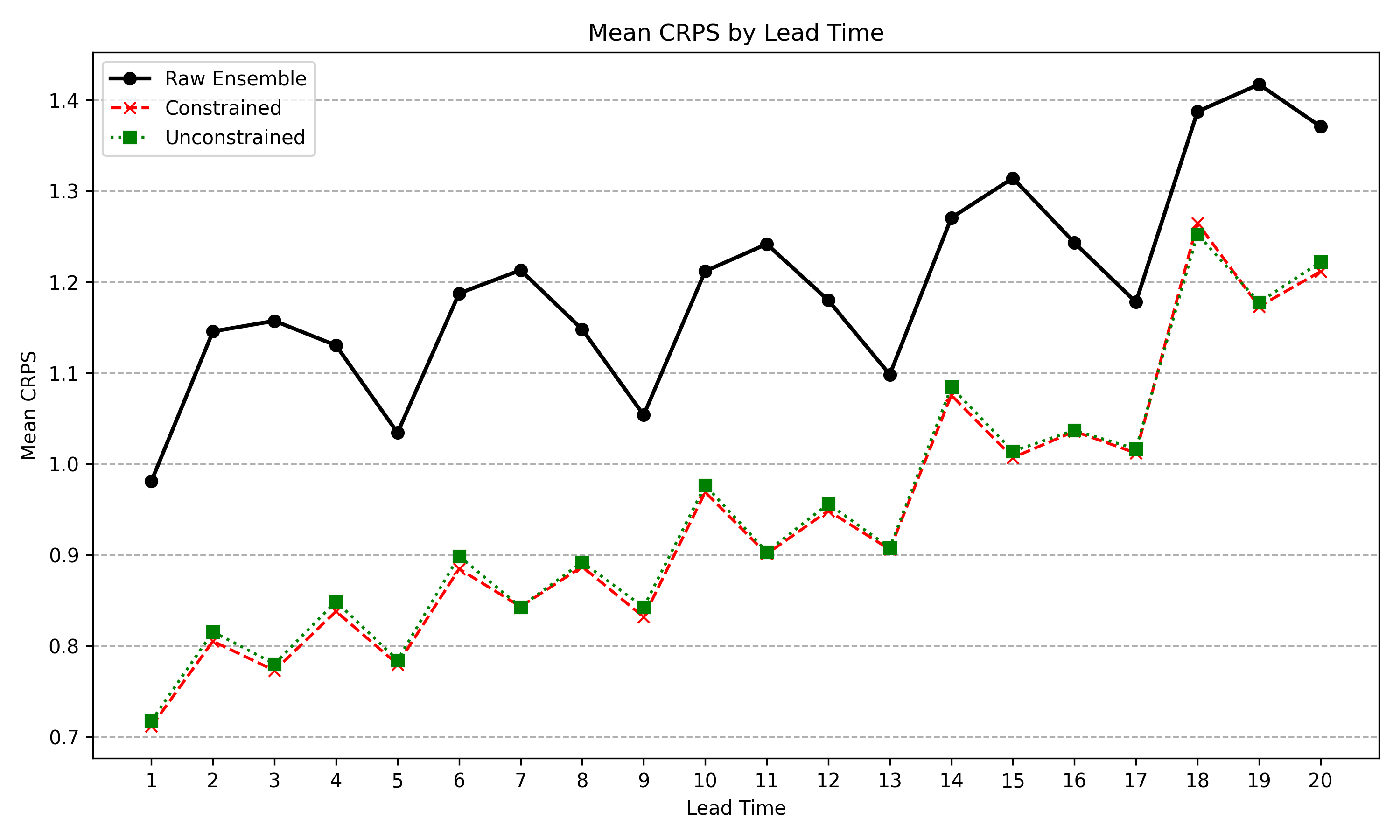}
    
    \caption{Average width of the nominal central prediction intervals and the mean CRPS values for the constrained and unconstrained V2 model on the test set}
    \label{fig:V2res}
\end{figure}

\begin{figure}[h]
    \centering
    \includegraphics[width=0.9\textwidth]{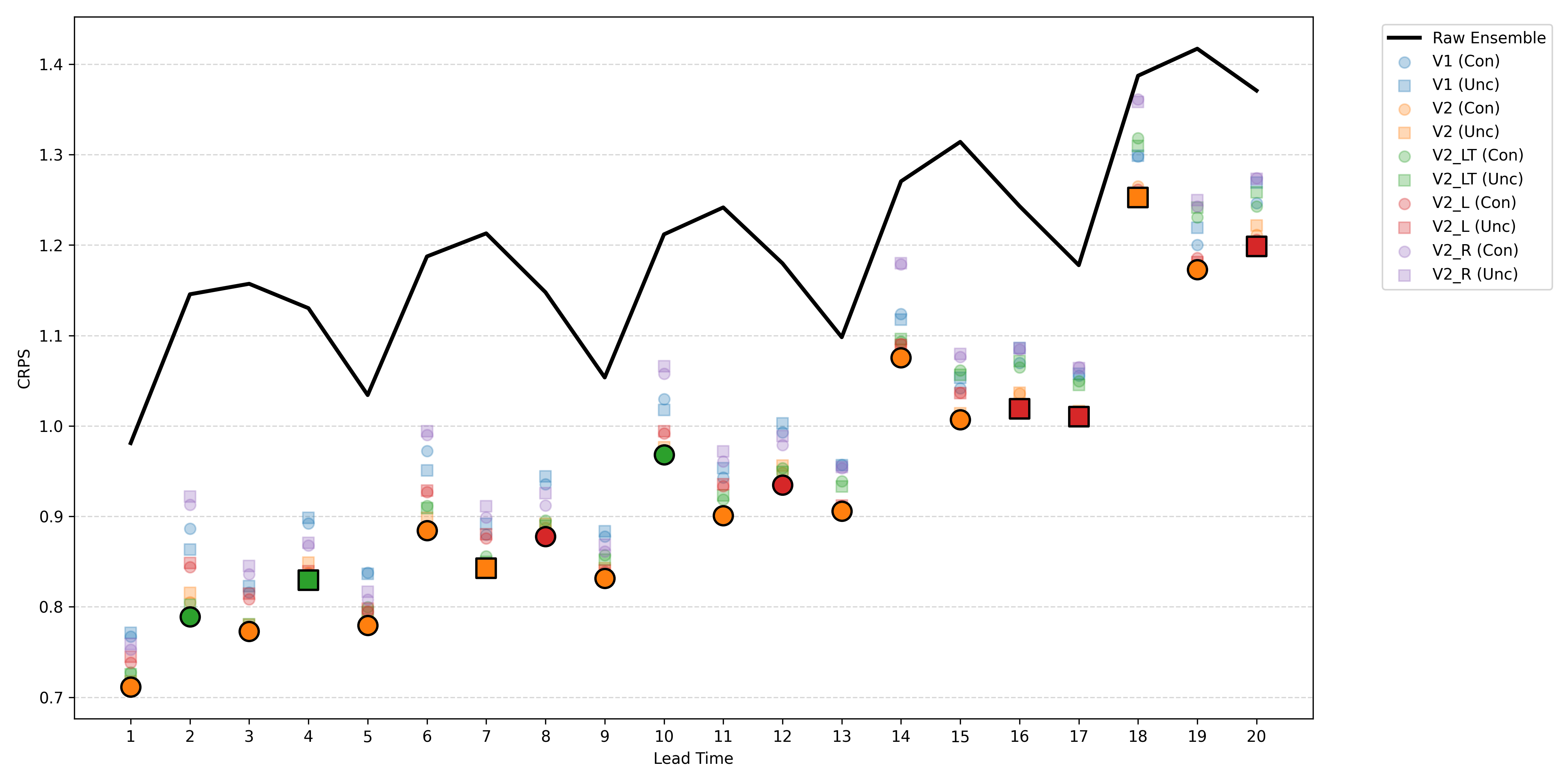} 
    \caption{CRPS as function of the lead time. The highlighted point denotes the best model in the given lead time. }
    \label{fig:crps_scatter}
\end{figure}

\begin{figure}[h!]
    \centering
    \includegraphics[width=0.95\textwidth]{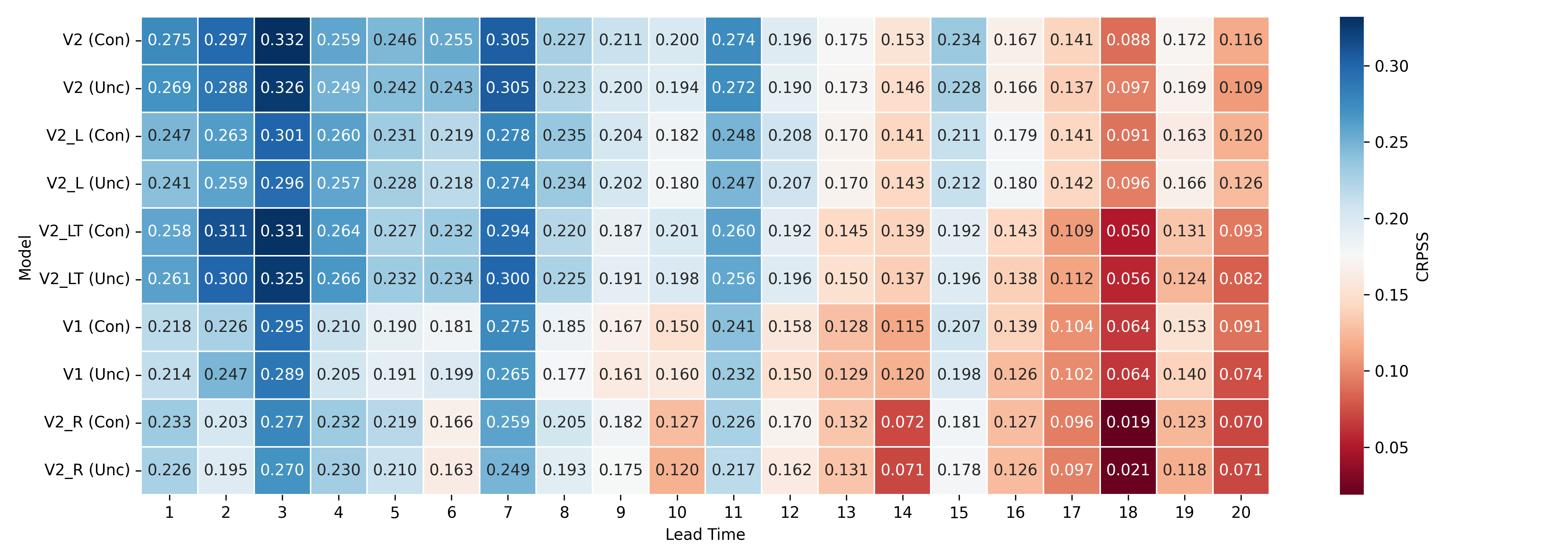}
    \caption{The heatmap visualization of the  CRPSS. The dark blue cells show the largest improvement compared to the raw ensemble. }
    \label{fig:crps_heatmap}
\end{figure}

The described changes are present at all lead times; the left part of Fig. \ref{fig:V2res} plots the average width of the nominal central prediction interval as a function of the lead time for the constrained and unconstrained versions of the V2 model, as well as for the raw ensemble, while  the right-hand side shows the mean CRPS values.   Instead of illustrating the phenomenon separately for each model, we show the summary of the lead time-wise changes in scatter plot and heatmap figures. In the first case, we plot the different evaluation metrics on the test set together with the values calculated from the raw ensemble. Each model's individual run is shown as a faint marker, while the result of the best-performing model in a given lead time is highlighted in solid color, enlarged in size, and with a black outline. The matrix-arranged heatmaps show the relative change; the rows  represent the different model configurations, and the columns represent the lead times. The values in the cells and the associated (typically red-blue) color scale make it possible to quickly visually decode the lead times in which a given model performs outstandingly or poorly.

Figure \ref{fig:crps_scatter}  plots the average CRPS values measured on the test as a function of the lead time, while Figure \ref{fig:crps_heatmap} shows the heatmap of the CRPSS value defined by formula \eqref{crpss} for the investigated models, evaluated on the test set, calculated by lead time (using the raw ensemble CRPS value as a reference). Since CRPSS is a positively oriented score, higher values represent a greater improvement. In the heatmap, the dark blue color indicates a larger improvement, while the red shades represent values close to the baseline. Fig. \ref{fig:crps_scatter} shows the dominance of the constrained models among the best performing models, even in the lead time-wise analysis, especially for short lead times. It can be observed, that the improvement of the CRPS is more significant closer to the initialization of the forecasts. Moreover,  we would like to draw attention to the cyclic pattern of the values; the markers could be grouped in group of four, corresponding to one day, thus, the largest improvements are at 18:00, every day (lead times 3,7,11,15,19).

\begin{figure}[h!]
    \centering
    \includegraphics[width=0.9\textwidth]{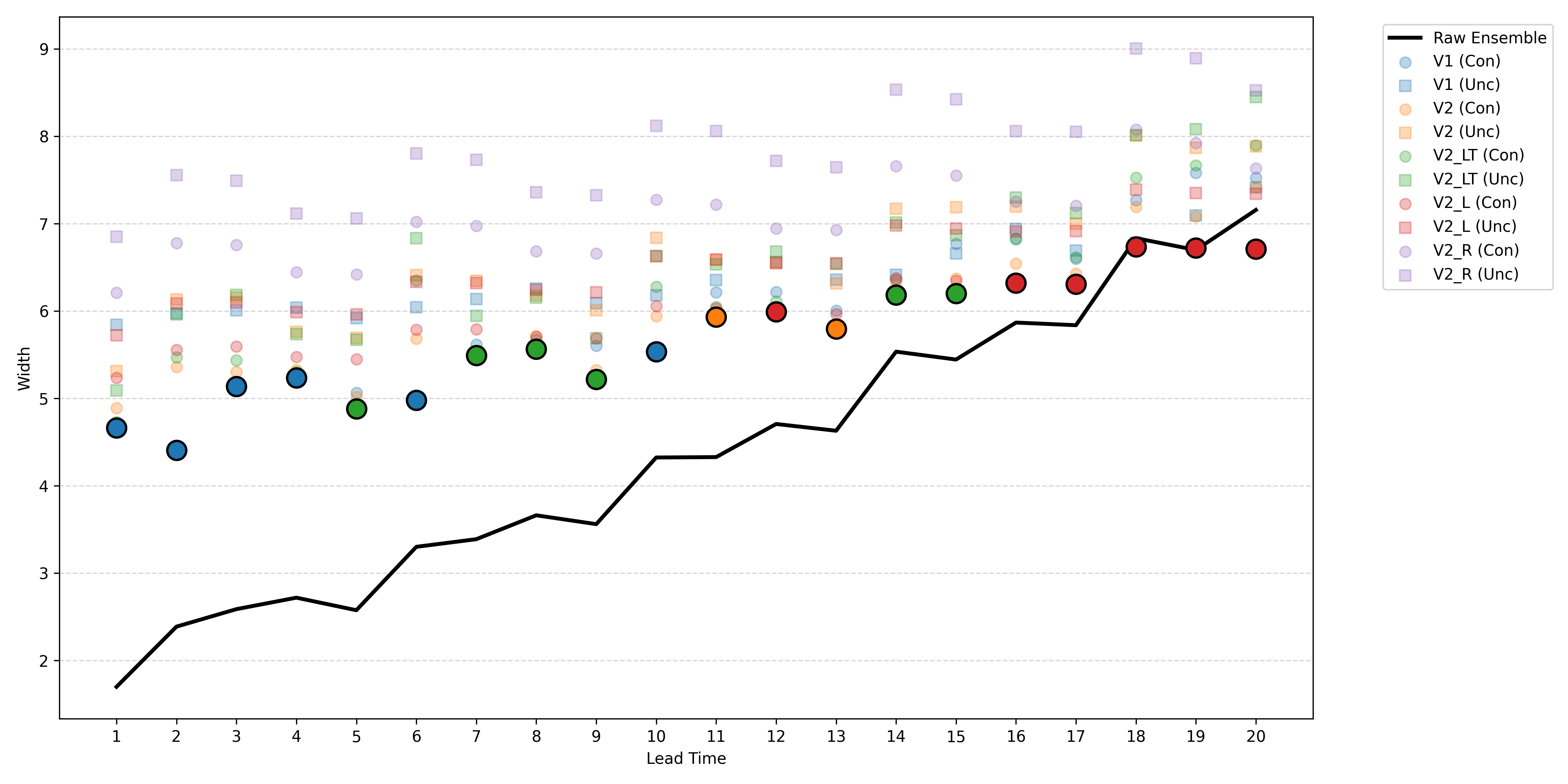} 
    \caption{The average width of the nominal central prediction intervals on the test set as a function of the lead time.
The highlighted points indicate the models that give the sharpest prediction (narrowest interval) for the given lead time.}
    \label{fig:width_scatter}
\end{figure}

\begin{figure}[h!]
    \centering
    \includegraphics[width=0.95\textwidth]{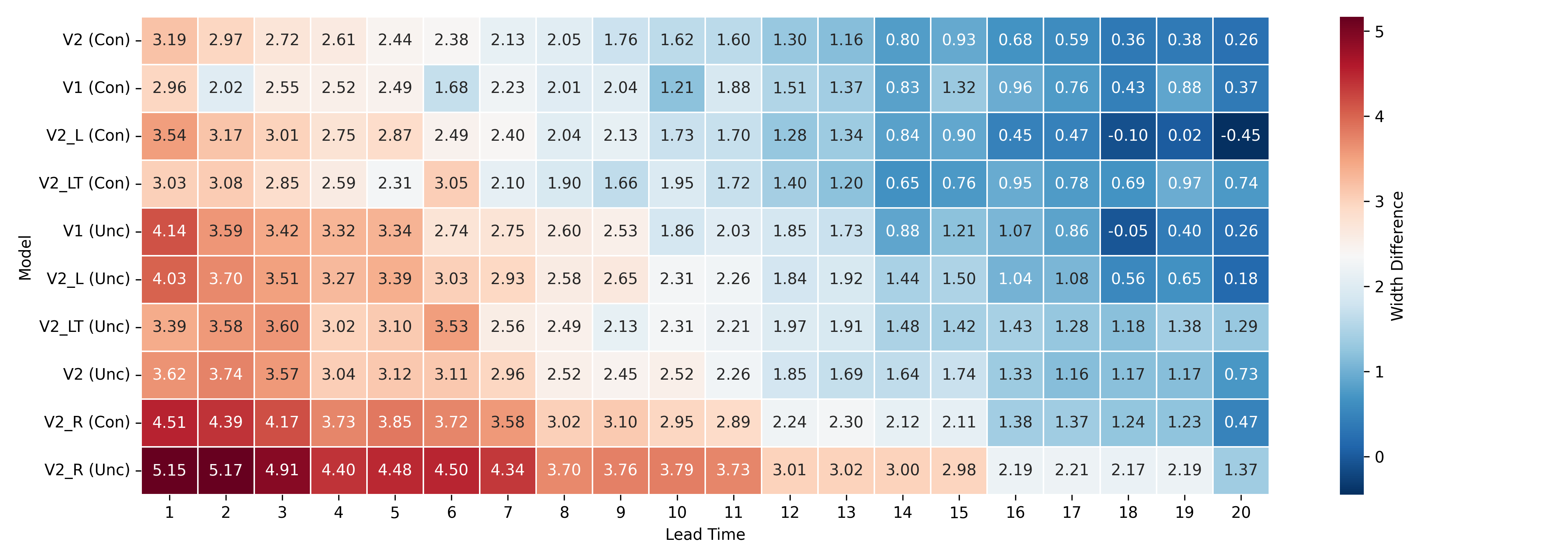}
    \caption{The heatmap of the width differences; the deviation of the average width obtained from the models from the average width calculated from the raw ensemble.}
    \label{fig:width_heatmap}
\end{figure}

Figures \ref{fig:width_scatter} and \ref{fig:width_heatmap} illustrate how much narrower the nominal central prediction intervals are in the case of the distributions generated by neural networks than the intervals spanned by the raw ensemble. Here we note that the aim was not to reduce the width compared to the raw ensemble, since one of the reasons for the weaker prediction skill of the raw ensemble is exactly the too narrow prediction interval. We wanted to improve the sharpness of the distributions obtained during post-processing, which is shown by the relative position of the markers of the same color. We considered it was important to present the results in the case of width in a common figure: on the one hand, it allows for a comparison of the individual post-processing models, on the other hand, it is visible that in the case of the largest lead times, even the width of the raw ensemble was improved.

\begin{figure}[h!]
    \centering
    \includegraphics[width=0.7\textwidth]{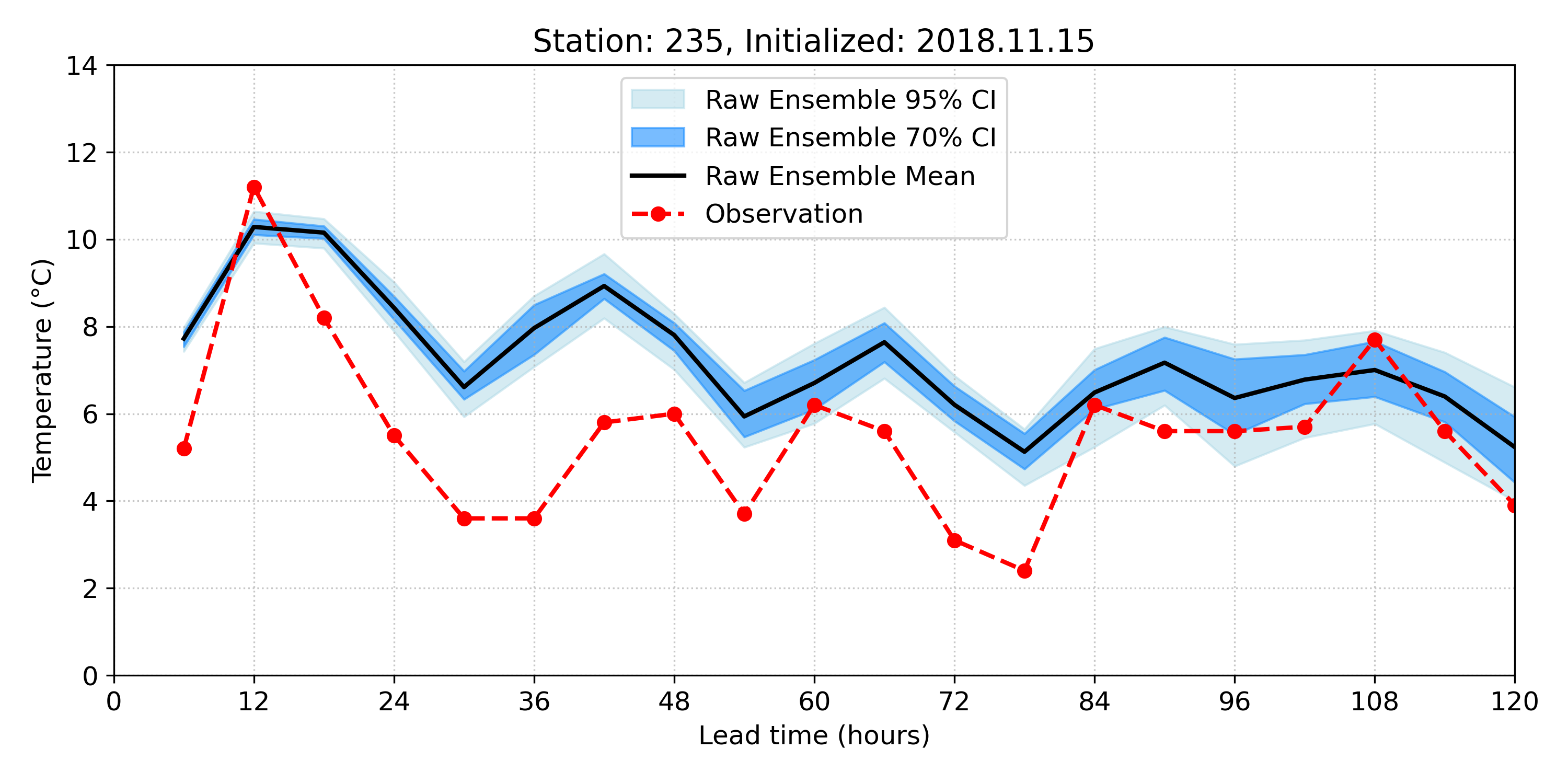}
    \caption{Confidence intervals of the raw ensemble forecasts for station De Kooy (Netherlands) initialized at 00:00 UTC, 15 November, 2018.}
    \label{fig:raw_235}
\end{figure}

\begin{figure}[h!]
    \centering
    \includegraphics[width=0.48\textwidth]{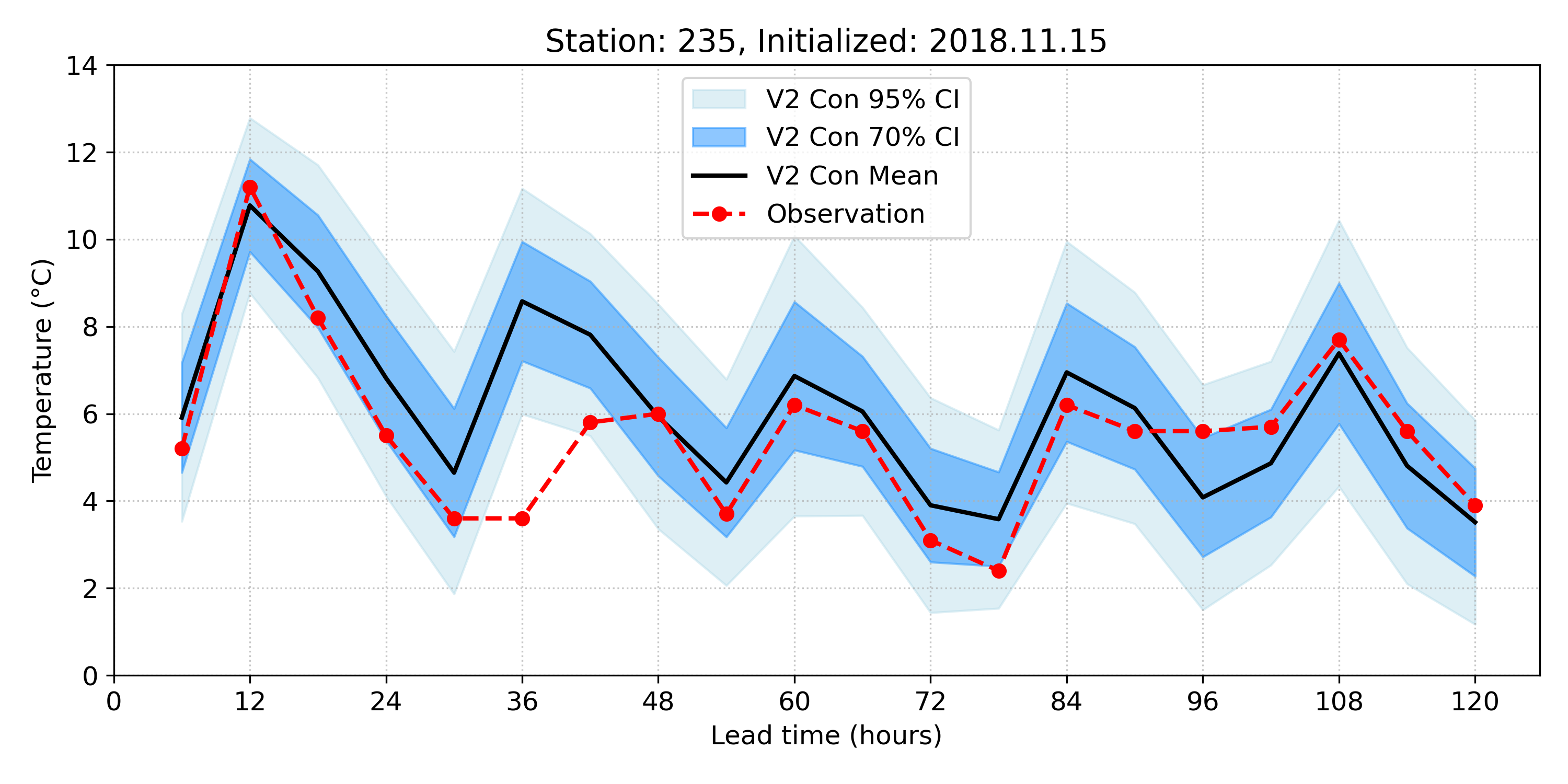}
    \hfill
    \includegraphics[width=0.48\textwidth]{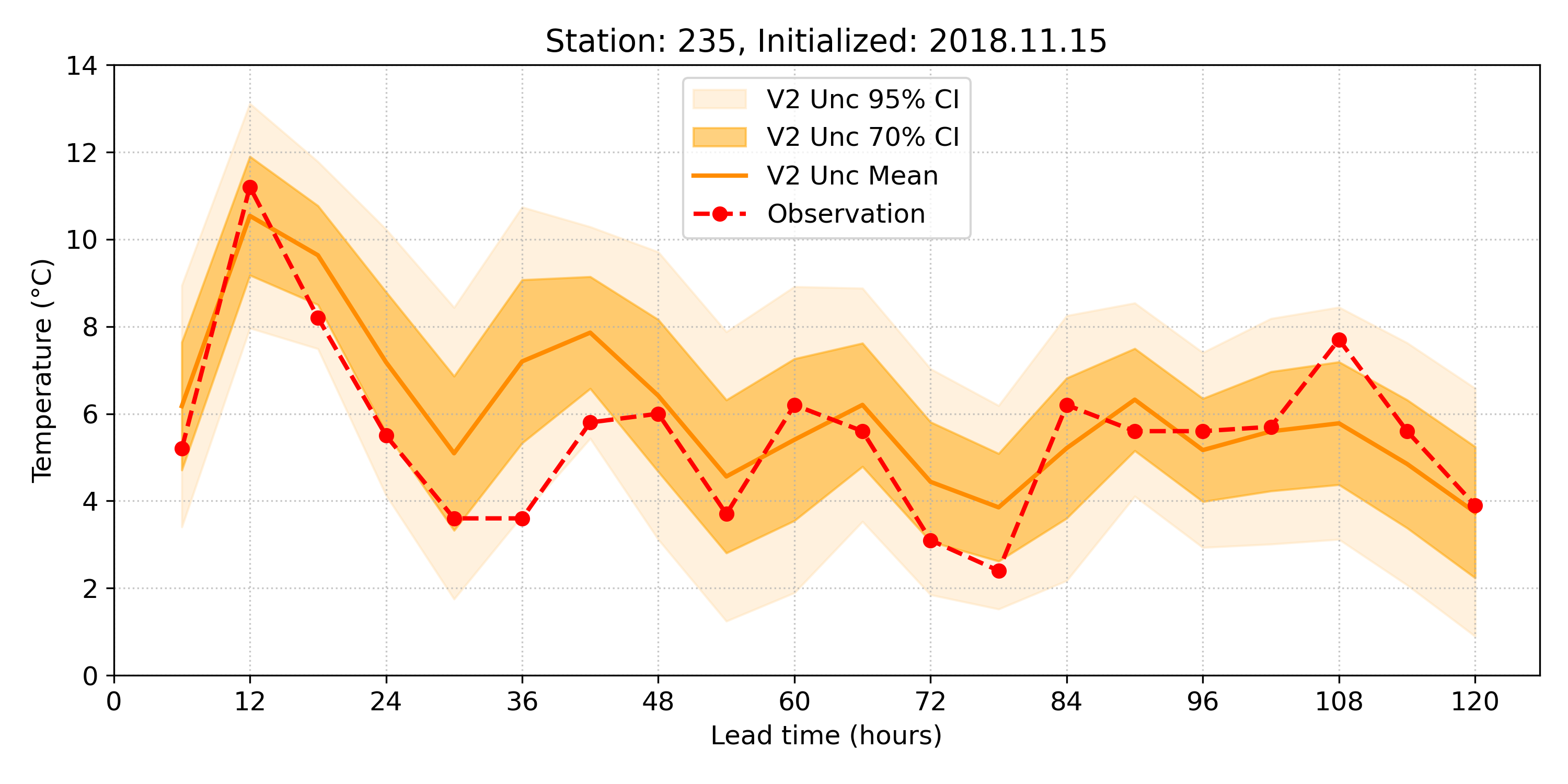}
    
    \caption{Comparison of the post-processed forecasts for the constrained and unconstrained V2 model (De Kooy (Netherlands) station, initialized at 00:00 UTC, 15 November, 2018).}
    \label{fig:global_235}
\end{figure}

To illustrate the effect of post-processing and the proposed penalty term, in addition to the previous figures showing average performance, we also plotted the $70\%$ and $95\%$ confidence intervals for the raw ensemble and the two post-processed cases for a specific station and initialization date, along with the validating observations. 
Fig. \ref{fig:raw_235} clearly shows the underdispersive and biased nature of the raw ensemble, strongly justifying the need for post-processing. In Fig. \ref{fig:global_235}  the confidence intervals are plotted for the constrained and unconstrained V2 model, together with the observations and with the mean of the predictive distribution, latter can be used as a point forecast at the given time horizon. 

\section{Conclusions}    

We propose a penalty technique to improve the sharpness of the predictive distribution in case of DRN-based parametric post-processing  methods. The loss function of the network, in this case the mean CRPS, is supplemented with an additional term, which is a scalar multiple of the width of predicted distribution's central prediction interval corresponding to a given confidence level determined by the number of the ensemble members. This width depends only on the variance of the distribution, but our assumption is that this penalization forces the optimizer to find a better located mean as well.  

We demonstrate the effect of the modified loss function in a case study on 2m temperature data for 122 stations in five European countries, downloaded from the EUPPBench dataset. Five models are applied differing in the network architectures, in the additional covariates used as input, and in the spatial and temporal resolution of the training.  All models are trained with two loss functions; without and with the application of the penalty term. The tests confirm the significant  decrease  in the widths when using the modified loss ($8.2\%-12.46\%$ relative change), while the mean CRPS on the test set does not deteriorate, and even a slight improvement can be observed. The same trend is visible in the RMSE scores, where the mean of the predictive distribution is used as a point forecast. We check the change in the distribution parameters; as an obvious consequence of the penalty, the standard deviations are smaller, but there is also a clear shift in the mean parameters.  

The  improvements of the previous evaluation metrics come with a slight decrease of the coverage values, but the extent is much smaller ($1.8\%-2.35\%$ relative change). A possible continuation of this work is to find a better hyperparameter setting which prevents or alleviates the negative change in the coverage values. 

We work with Gaussian distribution, as a typical predictive distribution for temperature forecast, however, as a next step, the presented method can be tested for other weather variables with different predictive distribution and CRPS. Based on the continuous improvement of the Ensemble Prediction System, and the introduction of the AIFS ensemble forecasts, further directions could be to validate the proposed method on a more recent dataset, as well as to  compare the effect in the case of IFS and AIFS ensemble forecasts.

\end{document}